%% file: main.tex
\documentclass[10pt,twocolumn,letterpaper]{article}

\usepackage{iccv}              


\definecolor{iccvblue}{rgb}{0.21,0.49,0.74}
\usepackage[pagebackref,breaklinks,colorlinks,allcolors=iccvblue]{hyperref}

\usepackage{algorithm}
\usepackage{algorithmic}
\usepackage{amsmath}
\def\paperID{12025} 
\def\confName{ICCV}
\def\confYear{2025}

\title{InfiniDreamer: Arbitrarily Long Human Motion Generation via \\
Segment Score Distillation}
\author{
Wenjie Zhuo$^{1, 2}$\quad Fan Ma$^{2}$\quad Hehe Fan$^{2\dagger}$\\
$^{1}$State Key Laboratory of Brain-Machine Intelligence, Zhejiang University \\
$^{2}$ReLER Lab, CCAI, Zhejiang University\\
{\tt\small \{wenjiezhuo, fanma, hehefan\}@zju.edu.cn}
}

\begin{document}

\twocolumn[{%
\renewcommand\twocolumn[1][]{#1}%
\maketitle
\thispagestyle{empty}
\begin{center}
    \vspace{-10mm}
    \includegraphics[width=0.98\textwidth]{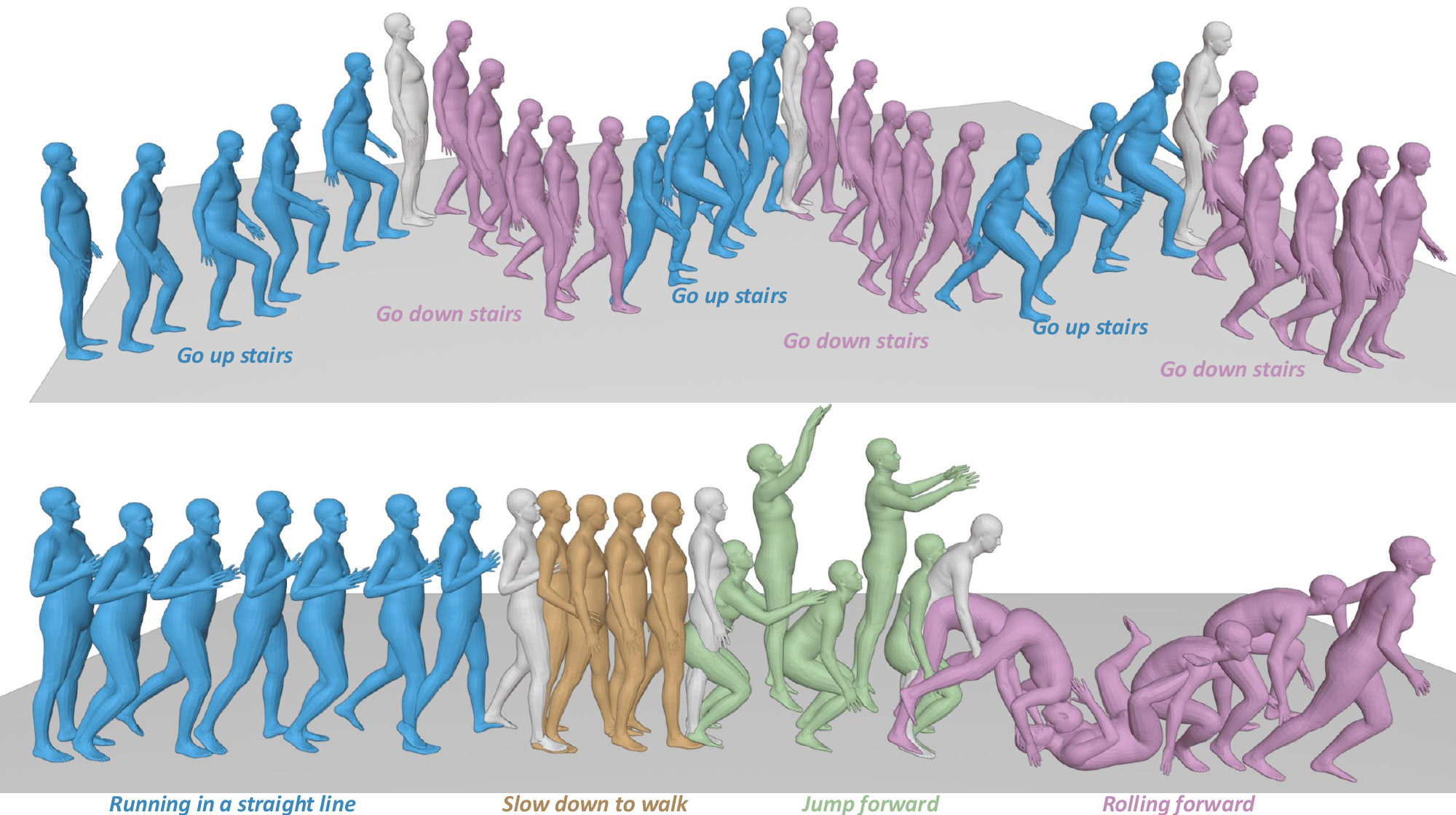}
    \vspace{-2mm}
    \captionof{figure}{\textbf{Long motion sequences generated by InfiniDreamer}. Given a list of textual prompts, our framework generates realistic human motions corresponding to each prompt, along with smooth and coherent transitions between them. This approach ultimately synthesizes a continuous, fluid long-duration human motion without requiring any long sequence data.}  
    \label{fig:teaser}
\end{center}%
}]
\renewcommand{\thefootnote}{\fnsymbol{footnote}}
\footnotetext{†Corresponding author.}
\input{sec/0_abstract}    
\input{sec/1_intro}
\input{sec/2_relate_work}
\input{sec/3_method}
\input{sec/4_experiments}
\input{sec/5_conclusion}
{
    \small
    \bibliographystyle{ieeenat_fullname}
    \bibliography{main}
}

\input{sec/X_suppl}

\end{document}

%% file: sec/0_abstract.tex
\begin{abstract}

We present InfiniDreamer, a novel framework for generating human motions of arbitrary length. Existing methods typically produce only short sequences, limited by the scarcity of long-range motion data. To address this, InfiniDreamer first generates short sub-motions for each textual description, then coarsely assembles them into a long sequence using randomly initialized transition segments. To refine this coarse motion, we introduce Segment Score Distillation (SSD)---an optimization-based approach that leverages a pre-trained motion diffusion model trained solely on short clips. SSD iteratively refines overlapping short segments sampled from the full sequence, progressively aligning them with the pre-trained short motion prior. This procedure ensures local fidelity within each segment and global consistency across segments. Extensive experiments demonstrate that InfiniDreamer produces coherent, diverse, and context-aware long-range motions without requiring additional long-sequence training.


\end{abstract}


%% file: sec/1_intro.tex
\section{Introduction}
\label{sec:intro}
This paper focuses on arbitrarily long 3D human motion generation~\cite{zhang2020perpetual, athanasiou2022teach, shafir2023human, lee2023multiact, chi2024m2d2m, barquero2024seamless}. It is a challenging task in computer vision, with great potential to benefit various downstream applications, such as AR/VR and film production. Benefiting from advancements in deep generative models~\cite{ho2020denoising, song2020denoising, nichol2021improved, radford2021learning,quan2024psychometry, song2025insert,feng2024evolved,feng2024efficient,feng2024videountier,zhuo2023whitenedcse,shi2025generative,shi2024fonts,shi2025wordcon, suo2024knowledge,suo2025trial,suo2025long,li2024personalvideo,li2025magicid} and the availability of large text-to-motion datasets~\cite{plappert2016kit, mahmood2019amass, punnakkal2021babel, guo2020action2motion, guo2022generating}, text-to-motion generation has seen significant progress. Recent approaches~\cite{tevet2023human, zhang2022motiondiffuse, chen2023executing, zhang2023generating, jiang2024motiongpt, guo2024momask, ahuja2019language2pose, zhang2023remodiffuse, yuan2023physdiff, dabral2023mofusion, petrovich2022temos, miao2025autonomous} are capable of generating realistic and coherent short motion sequences, typically around 10 seconds in duration. However, in most real-world applications, such as long-duration animations in gaming and full-body motion capture~\cite{ling2020character}, much longer motion sequences, spanning minutes or even hours, are often required. This gap presents a significant barrier to the broader applicability of current methods and highlights the critical need for advancements in generating continuous long-duration motions.

The primary challenge in generating arbitrarily long-motion sequences lies in the limited availability of high-quality long-sequence data~\cite{li2024infinite,han2024amd,punnakkal2021babel, li2023sequential, qian2023breaking}. Most existing datasets predominantly consist of short sequences annotated with single actions~\cite{guo2020action2motion, punnakkal2021babel} or simple textual descriptions~\cite{guo2022generating, plappert2016kit, mahmood2019amass}, lacking the temporal depth needed for continuous long-motion generation. To overcome these limitations, many previous works adopted auto-regressive models~\cite{athanasiou2022teach, lee2023multiact, li2023sequential, qian2023breaking}, which generate motions step-by-step based on previously generated frames. However, this auto-regressive nature often leads to the accumulation of errors over time, resulting in issues such as motion drift, repetitive patterns, and discontinuities over long motion sequences. Alternatively, some works utilize the infilling capabilities of motion diffusion models~\cite{shafir2023human, yang2023synthesizing, zhang2023diffcollage}. In these methods, motion segments are generated based on individual textual descriptions, and transitions between segments are filled in through in-painting. However, due to the strong modifications applied at the boundaries of each segment, this approach often leads to conflicts between adjacent motions, causing abrupt transitions, distortions in movement, or even overwriting of previously generated content.

To mitigate the issues, we turn to a smoother synthesis approach based on score distillation. Originally introduced by DreamFusion~\cite{poole2022dreamfusion}, Score Distillation Sampling (SDS)~\cite{poole2022dreamfusion} enables the creation of 3D assets using only a pre-trained text-to-image diffusion model. Unlike traditional diffusion sampling methods, which can result in abrupt local modifications, SDS~\cite{poole2022dreamfusion} emphasizes a gradual and smooth distillation process that maintains coherence across different views. Extending this advantage to temporal generation opens new possibilities for producing coherent long-duration human motion.

In this paper, we propose \textbf{InfiniDreamer}, a novel framework for generating arbitrarily long motion sequences. By contextually fine-tuning each sub-motion and refining transition segments between sub-motions, InfiniDreamer can generate coherent long-sequence motion in a training-free manner. Specifically, we first generate each sub-motion conditioned on its corresponding textual prompt and then assemble them into a coarsely extended sequence using randomly initialized transition segments. Next, we utilize a sliding window approach to iteratively sample short overlapping sequence segments from the coarse long motion sequence. We then propose \textbf{Segment Score Distillation (SSD)}, an optimization method that refines each short sequence segment by aligning it with the pre-trained motion prior. This segment-wise optimization ensures the local coherence of each sampled segment, while the refined transition segments maintain global consistency across the entire long motion sequence. After multiple rounds of optimization, our framework eventually yields coherent, contextually-aware long motion sequences. To verify the effectiveness of our framework, we evaluated the motion sequence and transition segments on two commonly used datasets, HumanML3D~\cite{guo2022generating} and BABEL~\cite{punnakkal2021babel}. The experimental results show that our method is significantly better than the previous training-free method. We also demonstrate our framework qualitatively, and the results show that our method has great contextual understanding capabilities, enable a seamless, coherent synthesis of long-duration motion sequences.

Overall, our contributions can be summarized as follows:

(1) We introduce InfiniDreamer, a novel framework capable of generating arbitrarily long human motion sequences in a training-free manner.

(2) We propose Segment Score Distillation (SSD), which iteratively refines overlapping short segments sampled from the coarse long motion. This process aligns each segment with the pre-trained motion prior, ensuring local and global consistency across the entire motion sequence.

(3) We conduct qualitative and quantitative evaluations of our framework. Experimental results show that our framework brings consistent improvement over the previous training-free methods.

%% file: sec/2_relate_work.tex
\section{Related Work}
\subsection{Text-to-Motion Generation}

Text-to-motion generation~\cite{zhang2022motiondiffuse, tevet2023human, zhang2023generating, tevet2022motionclip, petrovich2022temos, chen2023executing, dabral2023mofusion, yuan2023physdiff, zhang2023remodiffuse, jiang2024motiongpt, guo2024momask, ahuja2019language2pose, zhong2023attt2m, zhang2023finemogen}, which aims to create realistic human motions from textual descriptions, has gained substantial attention in recent years. Current works can be categorized into two main types: (i) GPT-based model and (ii) Diffusion-based Model. The former includes notable work such as T2M-GPT~\cite{zhang2023generating}, which combines VQ-VAE~\cite{van2017neural} with a transformer architecture for human motion generation from text, achieving impressive results. MotionGPT~\cite{jiang2024motiongpt} treats human motion as a foreign language and trains on a mixed motion-language dataset to build a unified motion-language model. MoMask~\cite{guo2024momask} proposes a masked transformer framework with residual transformer, enhancing text-to-motion generation. On the other hand, diffusion-based models are first introduced by MotionDiffuse~\cite{zhang2022motiondiffuse} and MDM~\cite{tevet2023human}. They developed a transformer-based diffusion model for generating motion based on text input. Rather than directly mapping raw motion data to text, MLD~\cite{chen2023executing} encodes motion into a latent space, improving the model's efficiency. Recently, Motion Mamba~\cite{zhang2024motion} combines state-space models (SSMs)~\cite{gu2023mamba} with diffusion models, offering an efficient framework for text-to-motion generation. All of these methods are capable of generating realistic and coherent human motion sequences, yet producing arbitrarily long human motion remains a challenge.

\input{fig/framework}

\subsection{Long Human Motion Generation}
Long human motion generation~\cite{li2024infinite,han2024amd,lee2024t2lm,chi2024m2d2m,shafir2023human,zhang2024motion,yang2023synthesizing,lee2023multiact,zhang2024infinimotion,qing2023story,li2023sequential,qian2023breaking,athanasiou2022teach,chen2025longanimation} is essential for many practical applications but remains constrained by limited datasets. Previous methods like Multi-Act~\cite{lee2023multiact} and TEACH~\cite{athanasiou2022teach} utilize a recurrent generation framework, and generate motion conditioned on the previously generated motion segment and the corresponding text prompt. However, these models suffer from error accumulation over time, causing issues like motion drift, repetitive patterns, and even 'freezing' after several iterations. To overcome this limitation, PCMDM~\cite{yang2023synthesizing} introduces a past-conditioned diffusion model alongside a coherent sampling strategy for long human motion generation. PriorMDM~\cite{shafir2023human} proposes an innovative sequential composition approach, which generates extended motion by composing prompted intervals and their transitions. FlowMDM~\cite{barquero2024seamless} proposes Blended Positional Encodings for seamless human motion composition. Recently, M2D2M~\cite{chi2024m2d2m} introduces adaptive transition probabilities and a two-phase sampling strategy to produce smooth and realistic motion sequences. In this work, we introduce a score distillation method, which refines randomly sampled short segments by aligning them with a pre-trained motion diffusion prior. This process ultimately generates a coherent and smooth long motion sequence.

%% file: fig/framework.tex
\begin{figure*}[!ht]
  \centering
  \includegraphics[width=0.98\textwidth]{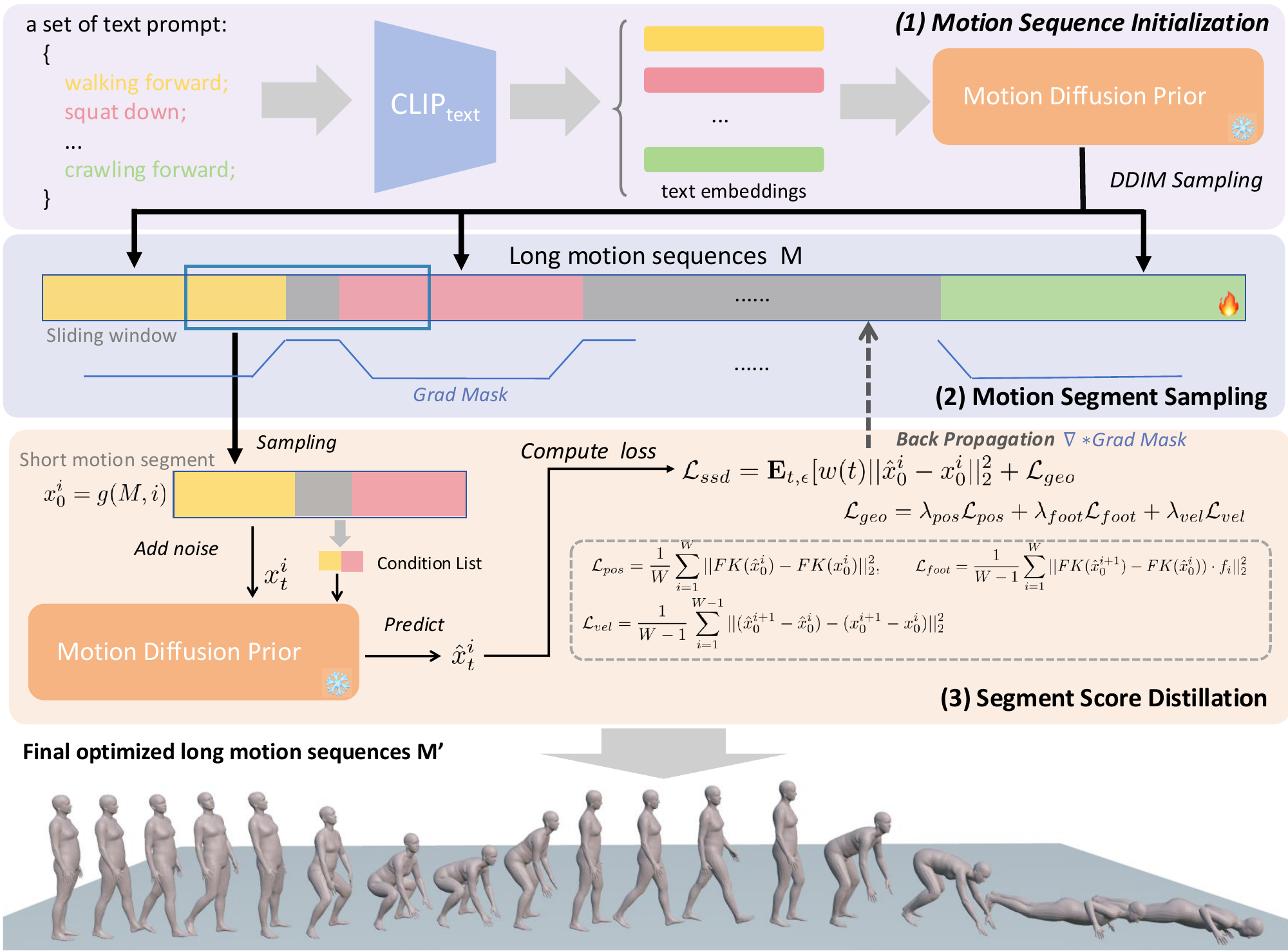}
  \vspace{-1mm}
  \caption{\textbf{Overview of InfiniDreamer} for arbitrarily long human motion generation. Given a list of text prompts, our framework generates a coherent and continuous long-sequence motion that aligns closely with each prompt. To achieve this, we start by initializing a long motion sequence using the (1) \textbf{Motion Sequence Initialization} module. Next, the (2) \textbf{Motion Segment Sampling} module iteratively samples short, overlapping sequence segments from the initialized motion. Finally, we refine each sampled segment with our proposed (3) \textbf{Segment Score Distillation}, optimizing each segment to align with the prior distribution of the pre-trained motion diffusion model. Through this iterative process, the framework synthesizes a seamless and fluid long-duration motion sequence, with realistic motions matching each prompt and smooth transitions connecting them.}  
\label{fig:overview}
\vspace{-2mm}
\end{figure*}

%% file: sec/3_method.tex
\section{Preliminaries}

\subsection{Score Distillation Sampling}
Score Distillation Sampling (SDS) was originally introduced in DreamFusion~\cite{poole2022dreamfusion} for the task of text-to-3D generation~\cite{poole2022dreamfusion,chen2023fantasia3d, huang2023dreamtime,wang2024prolificdreamer,chen2023textto3d,tang2023dreamgaussian, he2024customize,li2024gca}. It leverages the probability density distillation from a text-to-image diffusion model~\cite{rombach2022high} to optimize the parameters of any differentiable 3D generator, enabling zero-shot text-to-3D generation without requiring explicit 3D supervision. The flexibility of SDS allow it to guide various implicit representations like NeRF~\cite{mildenhall2020nerf}, 3DGS~\cite{kerbl3Dgaussians,zhuo2024vividdreamer} and image space~\cite{hertz2023delta,he2024freeedit,he2025plangen,DynamicID} towards high-fidelity results.

Formally, SDS utilizes a pre-trained diffusion prior $\phi$ to guide the implicit representation parameterized by $\theta$. Given a camera pose $\pi$, let $x=g(\theta; \pi)$ denote the image rendered from a differentiable rendering function $g$ with parameter $\theta$. SDS minimizes the density distillation loss between the posterior of $x=g(\theta; \pi)$ and the conditional density $p_\phi (x_t; y, t)$, which is a variational inference via minimizing KL divergence:
\begin{equation}
    \label{eq:distillation_loss}
    \begin{aligned}
        \min_{\theta} \mathcal{L}_{SDS}(\theta; x) = \mathbf{E}_{t,\epsilon}[\frac{\alpha_t}{\sigma_t} D_{KL}(q(x_t|g(\theta;\pi)||p_\phi (x_t; y, t)] ,
    \end{aligned}
\end{equation}

\noindent where $x_t=\alpha_t x+\sigma_t \epsilon$ with $\epsilon \sim \mathcal{N}(0, I)$. They further derive SDS by differentiating Eq.~\ref{eq:distillation_loss} with respect to differentiable generator parameter $\theta$, and omitting the U-Net Jacboian term to reduce computational cost and enhance performance. The SDS gradient update is given as follows:

\begin{equation}
    \label{eq:sds}
    \begin{aligned}
        \nabla_\theta \mathcal{L}_{SDS}(\theta;x)=\mathbf{E}_{t}[w(t)(\epsilon_\phi(x_t;y,t)-\epsilon)\frac{\partial x}{\partial \theta}] .
    \end{aligned}
\end{equation}

Previous works have demonstrated the effectiveness of score distillation loss in areas such as 3D generation~\cite{poole2022dreamfusion, wang2023score}, image editing~\cite{hertz2023delta} and video editing~\cite{zhu2024zero}. However, its application remains largely unexplored in other fields. In this work, we take a pioneering step by introducing Score distillation into the domain of long-sequence human motion generation, extending its utility to this challenging task.

\section{Methodology}


\subsection{Problem Definition}
Long-sequence human motion generation refers to the task of producing continuous and coherent motion sequences over extended time durations. Specifically, given a series of textual conditions $Y =\{y_1, y_2, y_3, ..., y_n\}$ as input, our goal is to generate the corresponding long-motion sequence $M = \{m_1, t_1, m_2, t_2, ..., m_n\}$, where $m_i$ represents the motion corresponding to each text prompts $y_i$, and $t_i$ denotes the transition segments between consecutive motion sequences $m_i$ and $m_{i+1}$. This task requires that each subsequence $m_i$ aligns closely with the corresponding textual condition $y_i$. In other words, the generated motion should accurately reflect the intent and meaning of each prompt. At the same time, each transition segment $t_i$ should feel as realistic and natural as possible. This ensures smooth transitions between the motion segments $m_i$ and $m_{i+1}$, allowing for a seamless flow in the overall motion sequence.

\subsection{InfiniDreamer}
As shown in Fig.~\ref{fig:overview}, Our proposed method, InfiniDreamer, consists of three main modules:

\noindent $\bullet$ \textbf{Motion Sequence Initialization Module}. The first stage in our framework is the initialization of the long motion sequence, serving as a foundational structure for further optimization. To create this initial sequence, we start by randomly initializing the entire long motion sequence $M$, which provides a rough, unsmoothed outline of the target motion. Then, we employ a pre-trained Motion Diffusion Model (MDM)~\cite{tevet2023human} to generate each motion segment $m_i$ within the sequence. Each segment $m_i$ is conditioned on the respective text prompt $y_i$, ensuring that the generated motion aligns semantically with the desired motion described in the prompt. 

Next, we introduce \textbf{gradient masks} to regulate the optimization process. Specifically, we assign a gradient mask $Mask_l$ to each motion segment and a different gradient mask $Mask_h$ to the transition regions. To ensure smooth transition between consecutive segments, we apply linear interpolation to the adjacent n frames between a motion segment and its neighboring transition region. This interpolation effectively bridges discontinuities and enhances the overall coherence of the generated sequence.

\noindent $\bullet$ \textbf{Motion Segment Sampling Module}. With the initialized sequence in place, we proceed to sample specific motion segments, which are essential for guiding the iterative optimization in subsequent steps. To achieve this, we employ a sliding window of size $W$, which moves along the long motion sequence with a stride size $S$. This sliding window technique allows us to iteratively sample overlapping short motion segments from the long sequence, denoted as $x^i_0$. By maintaining overlap between adjacent segments, the sliding window preserves continuity and smoothness between them, thereby enhancing the temporal coherence of the generated long motion sequence. For the text conditions used during optimization, we adopt the following strategy:

\begin{equation}
    \label{eq:text_cond}
    \begin{aligned}
    P(y = y_j) = 
    \begin{cases}
        1, & \text{if } x^i_0 \subseteq \text{sub-motion} j \\
        \frac{1}{n}, & \text{if } x^i_0 \text{ spans } n \text{ sub-motions}
    \end{cases}
    \end{aligned}
\end{equation}

\noindent where $y$ represents the text condition used in the optimization process, and $y_j$ denotes the text condition corresponding to the $j$-th sub-motion. This strategy ensures that the sampled motion segments retain meaningful textual associations, providing effective guidance during optimization.

\noindent $\bullet$ \textbf{Segment Score Distillation}. This module leverages a pre-trained motion diffusion model $\phi$ to optimize the distribution of the sampled short sequences, ensuring that each segment aligns with the underlying diffusion sample distribution. Specifically, Segment Score Distillation (SSD) iteratively optimizes each short motion segment to bring it closer to the high-quality distribution learned by the diffusion model, thereby enhancing the coherence and quality of the overall long motion sequence. To achieve this, for each sampled short motion segment $x^i_0$, we first randomly sample a timestep $t \sim \mathcal{U}(20, 980)$, then obtain each noised segment $x^i_t$ through $x^i_t=\sqrt{\alpha_t}x^i_0 + \sqrt{\sigma_t}\epsilon$, where $\alpha_t$ and $\sigma_t$ are noise scheduling parameters, and $\epsilon$ represents Gaussian noise. 

Using the motion diffusion model in an unconditional setting, we then incorporate an alignment loss to align the sampled motion segment with the predicted signal $\hat{x}^i_0 = \phi(x^i_t; t, \varnothing)$:
\vspace{-1mm}
\begin{equation}
    \label{eq:align_loss}
    \begin{aligned}
        \mathcal{L}_{align} = \mathbf{E}_{t, \epsilon}[w(t)||\hat{x}^i_0 - x^i_0||^2_2], 
    \end{aligned}
\end{equation}

\noindent where $w(t)$ is weighting function. To further improve the coherence and realism of generated motions, we augment our method with three commonly used geometric losses, inspired by prior work~\cite{petrovich2021action,shi2020motionet,tevet2023human}. These include (1) positional constraints when predicting rotations, (2) foot contact constraints to maintain stable ground interaction, and (3) velocity regularization to encourage smooth transitions: 
\vspace{-1mm}
\begin{equation}
    \label{eq:pos_loss}
    \begin{aligned}
        \mathcal{L}_{pos} = \frac{1}{W}\sum_{i=1}^W|| FK(\hat{x}^i_0) - FK(x^i_0)||^2_2 ,
    \end{aligned}
\end{equation}
\vspace{-1.mm}
\begin{equation}
    \label{eq:foot_loss}
    \begin{aligned}
        \mathcal{L}_{foot} = \frac{1}{W-1}\sum_{i=1}^{W}|| FK(\hat{x}^{i+1}_0) - FK(\hat{x}^i_0))\cdot f_i||^2_2 ,
    \end{aligned}
\end{equation}
\vspace{-1.mm}
\begin{equation}
    \label{eq:vel_loss}
    \begin{aligned}
        \mathcal{L}_{vel} = \frac{1}{W-1}\sum_{i=1}^{W-1}||(\hat{x}^{i+1}_0 - \hat{x}^i_0) - (x_0^{i+1} - x_0^{i}) ||^2_2 ,
    \end{aligned}
\end{equation}

\noindent where $FK(\cdot)$ is the forward kinematic function, it maps joint rotations to joint positions (or acts as the identity function if joint rotations are not predicted). $f_i \in {0, 1}^J$ is the binary foot contact mask for each frame $i$, it denotes whether the foot touches the ground, helping mitigate foot-sliding by nullifying ground-contact velocities. Together, our final Segment Score Distillation loss is:
\vspace{-1mm}
\begin{equation}
    \label{eq:slide_score_distillation}
    \begin{aligned}
        \mathcal{L}_{ssd} = \mathcal{L}_{align} + \lambda_{pos}\mathcal{L}_{pos} + \lambda_{foot}\mathcal{L}_{foot} + \lambda_{vel}\mathcal{L}_{vel} ,
    \end{aligned}
\end{equation}


\noindent where $\lambda_{pos}$, $\lambda_{foot}$ and $\lambda_{vel}$ are hyper-parameters that balance the contribution of each geometric loss in the overall objective function. We set them to $0$ in our experiments on the HumanML3D dataset and set them to $0.1$ for the BABEL dataset.

%% file: sec/4_experiments.tex
\section{Experiments}
\input{table/humanml3d_eval}

\input{table/babel_eval}

\subsection{Datasets}
We evaluate InfiniDreamer on two datasets, HumanML3D~\cite{guo2022generating} and BABEL~\cite{punnakkal2021babel}, which are essential benchmarks for assessing motion generation models.

\noindent\textbf{HumanML3D}. The HumanML3D dataset~\cite{guo2022generating} consists of 14,616 motion samples, each paired with 3-4 textual descriptions, enabling the model to learn from rich, multi-perspective annotations. These motions are sampled at 20 FPS and derive from the AMASS~\cite{mahmood2019amass} and HumanAct~\cite{guo2020action2motion} datasets, with additional manual text descriptions for greater semantic detail. HumanML3D utilizes a 263-dimensional pose vector that encodes joint coordinates, angles, velocities, and feet contact information, allowing for precise motion modeling. For evaluation, we use motions with lengths ranging from 40 to 200 frames.

\noindent\textbf{BABEL}. The BABEL dataset~\cite{punnakkal2021babel} consists of 10,881 sequential motion samples and a total of 65,926 segments, wherein each segment correlates with a distinct textual annotation. This high level of segmentation supports the modeling of nuanced transitions and distinct action phases, making BABEL a valuable benchmark for evaluating long-motion generation. During evaluation, we follow the setting of PriorMDM~\cite{shafir2023human}, which refines BABEL by excluding poses like `a-pose' or `t-pose' and combines transitions with subsequent actions to create smoother sequences.

\subsection{Implementation details}
We use the Motion Diffusion Model (MDM)~\citep{tevet2023human} as our short motion prior, and set $Mask_l=0.1$, $Mask_h=0.8$ for all experiments. For HumanML3D, we use the pre-trained model with 0.6M steps trained on this dataset, while for BABEL, we use the pre-trained model with 1.25M steps. For the optimization process, we set the guidance scale as $7.5$ and sample time steps $t \sim \mathcal{U}(20, 980)$. For all results, we set the patch size as $120$ and the stride size as $30$. We optimized all long motion sequences for 20000 iterations using AdamW~\cite{loshchilov2019decoupled} optimizer with a learning rate $0.002$. We conduct all experiments on a single A6000 GPU.
\input{fig/compare}

\input{fig/abl_lr}

\subsection{Evaluation metrics}
To evaluate our approach, we utilize the following metrics: (1) R-Precision which measures the semantic alignment between the input text and the generated motions. R-Precision measures the degrees to which generated motions align with the provided textual descriptions, while Multimodal Distance evaluates the consistency between multiple generated motions and the text. (2) Frechet Inception Distance (FID), which evaluates the overall quality of motions by comparing the distribution of high-level features between generated motions and real motions. A lower FID indicates a closer resemblance to real motions. (3) Diversity measures the variability and richness of the generated motion sequences. (4) The Multimodal Distance metric, which measures the diversity of motions generated from a single text prompt, indicates the model's ability to generate varied interpretations of the same input.
\subsection{Quantitative Comparisons}
Following previous works~\cite{athanasiou2022teach, shafir2023human}, we quantitatively evaluate the quality of generated long motion sequences on HumanML3D~\cite{guo2022generating} and BABEL~\cite{punnakkal2021babel}. For motion sequences, we use R-precision, FID, Diversity, and Multimodal distance metrics to measure their quality, while for transition segments, we use FID and Diversity to measure their quality. As shown in Table~\ref{tab:humanml3d_eval} and Table~\ref{tab:babel_eval}, InfiniDreamer brings consistent improvement over the current state-of-the-art methods. In HumanML3D, our framework outperforms previous methods across all evaluation metrics. The generated sequences demonstrate a higher degree of alignment with the input text and closely match the distribution of real data. Additionally, our framework achieves superior results in generating transition segments. In Babel, our framework achieves a significant advantage in R-precision, indicating better alignment between the generated motions and the textual descriptions. Furthermore, when we apply the geometric loss, our model demonstrates additional improvements in the FID metric, enhancing the realism and quality of the generated motion sequences.
 

\subsection{Qualitative Comparisons}
To further showcase the advantages of our approach, we conduct qualitative experiments to compare our method with DoubleTake~\cite{shafir2023human}. We present two comparative experiments. In the upper row of Fig.~\ref{fig:compare}, we use two identical prompts, ``a main is jogging forward.'' DoubleTake generates a nearly frozen transitional motion between them, while our method produces a smoother transition that maintains global coherence. In the second row, DoubleTake suffers from motion loss, distortion, and mismatched motions---for example, it fails to generate the dodge motion and incorrectly produces a sit crisscrossed motion. In contrast, our method successfully avoids these problems. Both examples validate the superiority of our framework.
\input{table/abl_patch}

\section{Ablation study}
\noindent \textbf{Ablation on Sliding Window Size $W$.} In Tab.~\ref{tab:abl_patch}, we present the impact of the hyper-paraeter Sliding Window Size $W$ on model performance. $W$ controls the size of each sampled segment, whereas a larger $W$ allows the model to incorporate more contextual information. We observe that with a very small $W$, the performance of transition segments declines sharply. However, as $W$ increases, the transition quality exhibits fluctuating declines. This suggests that a moderate context length is beneficial for transition generation, whereas an overly extended context introduces interference. In terms of motion segment generation, performance consistently decreases as $W$ grows. We speculate this is due both to MDM's limitations in handling long sequences and to the interference in semantic alignment caused by excessive context length.

\noindent \textbf{Ablation on Stride Size $P$.} In Tab.~\ref{tab:abl_patch}, we also examine the impact of the hyper-parameter Stride Size $P$ on model performance. $P$ controls the frame shift of the sliding window in each step and, consequently, the overlap between segments. Our experiments show that when $P < W$, this parameter has minimal impact on performance. However, when $P \geq W$, there is a noticeable improvement in motion generation, whereas transition generation performance drops sharply.

The improvement in motion generation can be attributed to the nature of SSD, which adds noise during optimization. This noise slightly degrades motion quality. When $P \geq W$, certain motion frames are excluded from sampling and thus bypass SSD optimization, resulting in performance gains. In contrast, since transitions are initialized randomly, excluding certain transition frames from optimization effectively leaves them as random noise, which severely impacts transition quality.

\noindent \textbf{Ablation on Learning Rate $\eta$.} The learning rate $\eta$ controls the update magnitude for long motion sequence parameters. In Fig.~\ref{fig:abl_lr}, we illustrate the effect of different learning rates on motion sequence generation. We observe that an excessively high learning rate will cause the motion amplitude to gradually decrease, eventually resulting in a static output, leading to motion loss. As shown in the top row of Fig.~\ref{fig:abl_lr}, where a kicking motion is optimized into stillness. Conversely, a lower learning rate leads to under-training, introducing more noise and causing motion distortions. As shown in the bottom of Fig.~\ref{fig:abl_lr}, we notice significant motion distortion and exaggerated amplitude in the transitional phase preceding the kick motion, highlighting the need for a balanced learning rate.

%% file: table/humanml3d_eval.tex
\begin{table*}[!t]
\centering
\setlength{\tabcolsep}{4pt}
\renewcommand{\arraystretch}{1.2}
\resizebox{\textwidth}{!}{%
\begin{tabular}{lcccc||cc}
    \toprule 
    & \multicolumn{4}{c}{Motion} & \multicolumn{2}{c}{Transition (30 frames)} \\
    \hline 
    & R-precision $\uparrow$ & FID $\downarrow$ & Diversity $\rightarrow$ & MultiModal-Dist $\downarrow$ & FID $\downarrow$ & Diversity $\rightarrow$ \\
    \hline
    Ground Truth & $0.797 \pm 0.003$ & $1.6 \cdot 10^{-3} \pm 0.00$ & $9.59 \pm 0.13$ & $2.98 \pm 0.01$ & $1.8 \cdot 10^{-3} \pm 0.00$ & $9.55 \pm 0.09$ \\
    \hline 
    DoubleTake~\cite{shafir2023human} & $0.603 \pm 0.009$ & ${1.36 \pm 0.03}$ & $9.33 \pm 0.15$ & $4.27 \pm 0.01 $ &$\underline{3.19 \pm 0.29}$ & $\underline{8.09 \pm 0.08}$ \\ 
    DiffCollage~\cite{zhang2023diffcollage} &$\underline{0.605 \pm 0.006}$ & $\underline{1.07 \pm 0.05}$ & $\underline{9.34 \pm 0.11}$ & $\underline{3.62\pm0.01}$ & $4.27 \pm 0.09$ & $7.47 \pm 0.08$ \\
    
    \hline
    InfiniDreamer (ours) & $\mathbf{0.679 \pm 0.007}$ & $\mathbf{0.47 \pm 0.12}$ & $\mathbf{9.58 \pm 0.15} $ & $\mathbf{3.15 \pm 0.01}$ & $\mathbf{2.04\pm 0.28}$ & $\mathbf{8.69 \pm 0.09}$\\  
    \quad w/o gradient masks & $0.643 \pm 0.011$ & $0.64 \pm 0.09$ & $9.60 \pm 0.17 $ & $3.32 \pm 0.01$ & $2.25\pm 0.34$ & $8.51 \pm 0.10$\\  
    \bottomrule 
    
\end{tabular}%
}
\vspace{-2mm}
\caption{Comparison of InfiniDreamer with the state-of-the-art training-free methods in HumanML3D. Symbols $\uparrow$, $\downarrow$, and $\rightarrow$ mean that higher, lower, or closer to the ground truth (GT) value are better, respectively. We run each evaluation 10 times to obtain the final results. We use \textbf{Bold} to indicate the best result, and use \underline{underline} to indicate the second-best result.  }
\label{tab:humanml3d_eval}
\end{table*}

%% file: table/babel_eval.tex
\begin{table*}[!t]
\centering
\setlength{\tabcolsep}{4pt}
\renewcommand{\arraystretch}{1.2}
\resizebox{\textwidth}{!}{%
\begin{tabular}{lcccc||cc}
    \toprule 
    & \multicolumn{4}{c}{Motion} & \multicolumn{2}{c}{Transition (30 frames)} \\
    \hline 
    & R-precision $\uparrow$ & FID $\downarrow$ & Diversity $\rightarrow$ & MultiModal-Dist $\downarrow$ & FID $\downarrow$ & Diversity $\rightarrow$ \\
    \hline
    Ground Truth & $0.629 \pm 0.001$ & $0.4 \cdot 10^{-3} \pm 0.00$ & $8.52 \pm 0.09$ & $3.51 \pm 0.01$ & $0.7 \cdot 10^{-3} \pm 0.00$ & $8.23 \pm 0.14$ \\
    \hline 
    TEACH~\cite{athanasiou2022teach} & $0.461 \pm 0.012$ & $1.43 \pm 0.04$ & $7.71 \pm 0.11$ & $7.93 \pm 0.01$  & $4.23 \pm 0.37$ & $\mathbf{8.37 \pm 0.11}$\\
    DoubleTake~\cite{shafir2023human} & $0.483 \pm 0.009$ & ${1.14 \pm 0.05}$ & $\underline{8.28 \pm 0.09}$ & $6.97 \pm 0.01 $ &$3.54 \pm 0.10$ & $7.31 \pm 0.12$ \\ 
    DiffCollage~\cite{zhang2023diffcollage} &$0.487 \pm 0.009$ & $1.83 \pm 0.05$ & $7.89 \pm 0.11$ & $6.74\pm 0.01$ & $4.62 \pm 0.09$ & $7.07 \pm 0.08$ \\

    \hline
    InfiniDreamer (ours) & $\mathbf{0.543 \pm 0.009}$ & $\mathbf{0.97 \pm 0.09}$ & $\mathbf{8.31 \pm 0.06} $ & $\mathbf{5.80 \pm 0.01}$ & $\mathbf{2.07\pm 0.30}$ & $\underline{7.95 \pm 0.07}$\\
    \quad w/o geo losses & $\underline{0.537 \pm 0.008}$ & $\underline{1.09 \pm 0.09}$ & $\underline{8.28 \pm 0.06} $ & $\underline{5.87 \pm 0.01}$ & $\underline{2.15\pm 0.31}$ & $7.84 \pm 0.09$\\
    \quad w/o gradient masks & ${0.522 \pm 0.008}$ & ${1.14 \pm 0.09}$ & ${8.17 \pm 0.07} $ & ${6.12 \pm 0.01}$ & ${2.37\pm 0.27}$ & $7.72 \pm 0.07$\\
    \bottomrule 
    
\end{tabular}%
}
\vspace{-2mm}
\caption{Comparison of InfiniDreamer with the state-of-the-art training-free methods in BABEL. Symbols $\uparrow$, $\downarrow$, and $\rightarrow$ mean that higher, lower, or closer to the ground truth (GT) value are better, respectively. We run each evaluation 10 times to obtain the final results. We use \textbf{Bold} to indicate the best result, and use \underline{underline} to indicate the second-best result. }
\label{tab:babel_eval}
\end{table*}

%% file: fig/compare.tex
\begin{figure*}[!ht]
  \centering
  \includegraphics[width=0.98\textwidth]{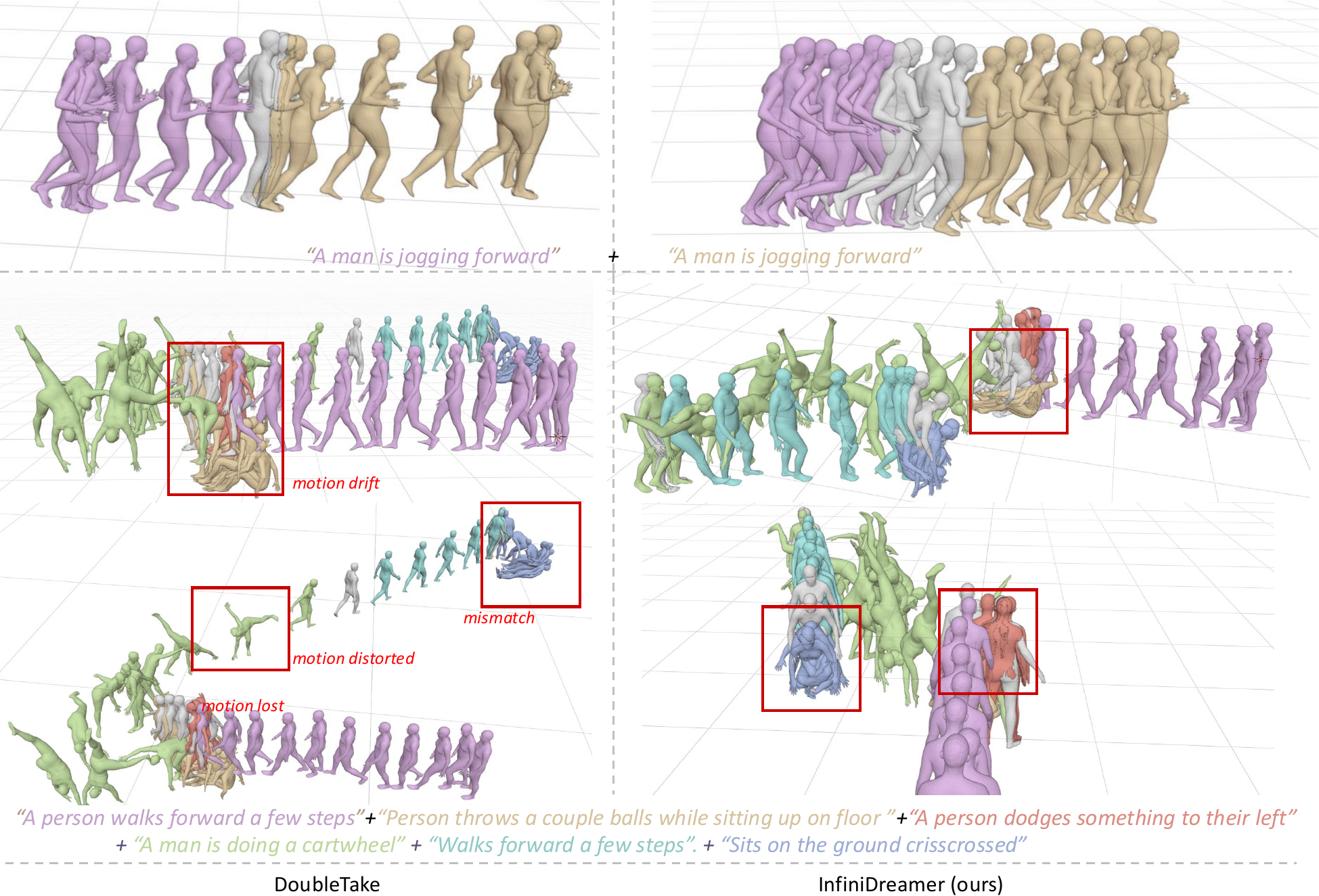}
  \vspace{-1mm}
  \caption{Qualitative Comparisons to Baseline for Long Motion Generation. We present two examples: in the top row, our framework demonstrates strong segment transition capabilities, effectively generating a smooth jogging transition between two jogging motions. In contrast, the baseline produces a transitional segment with noticeable pauses. In the second row, we test a more complex and fine-grained example. The baseline method generates drifting motions, misses the segment ``dodges something to their left'', and introduces mismatched motion such as ``crisscrossing''. In comparison, our method produces a higher-quality sequence with enhanced fine-grained comprehension.}
\label{fig:compare}
\end{figure*}


%% file: fig/abl_lr.tex
\begin{figure}[!ht]
  \centering
  \includegraphics[width=\columnwidth]{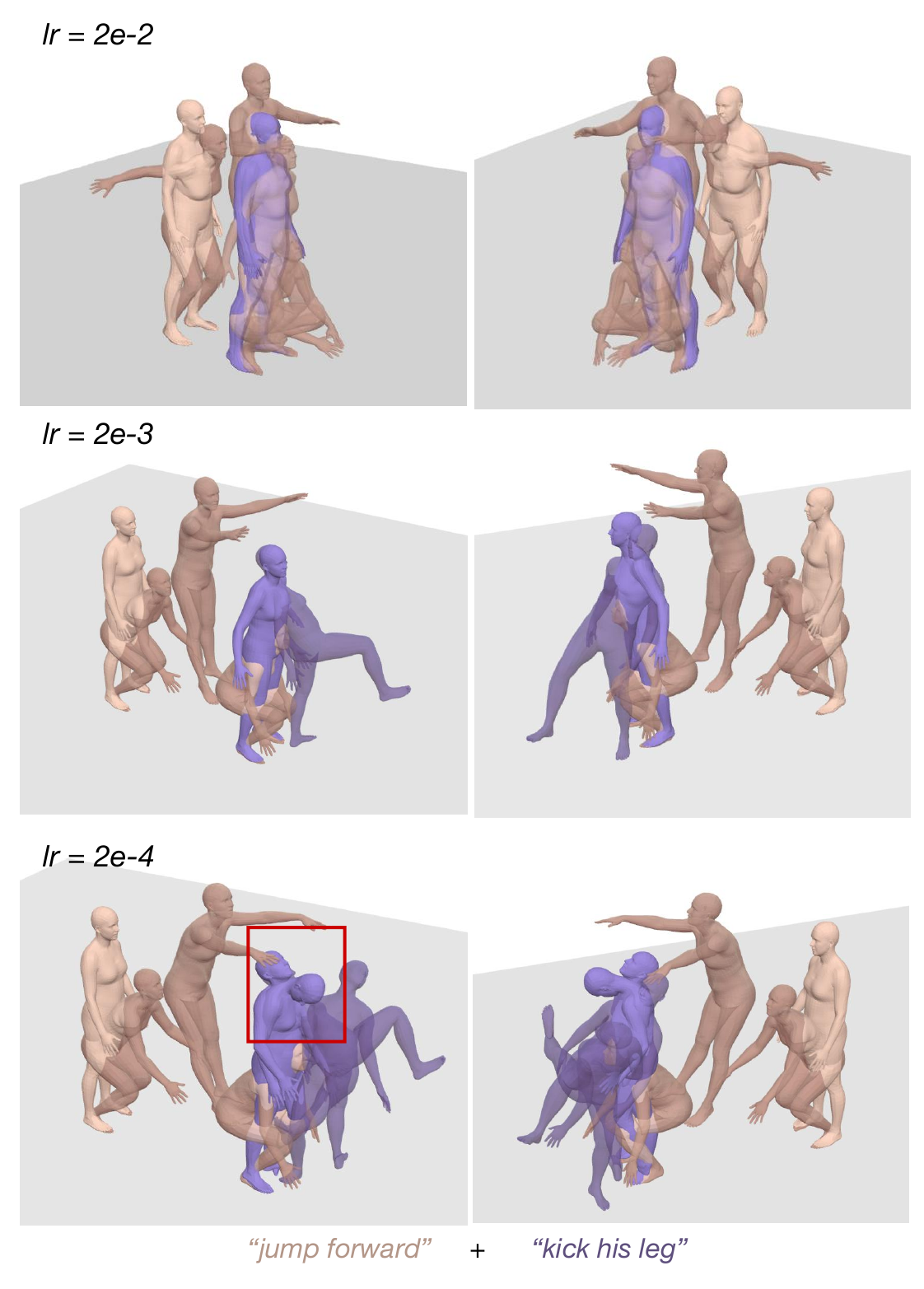}
  \vspace{-7mm}
  \caption{Ablation Study on Learning Rate $\eta$. We experiment with different $\eta$ and find that an excessively high learning rate leads to motion stillness (\ie, motion lost), while a lower learning rate results in large noise disturbances, causing motion distortions.}
\label{fig:abl_lr}
\end{figure}
\vspace{-2mm}

%% file: table/abl_patch.tex
\begin{table*}[!t]
\centering
\setlength{\tabcolsep}{4pt}
\renewcommand{\arraystretch}{1.2}
\resizebox{\textwidth}{!}{%
\begin{tabular}{lcccc||cc}
    \toprule 
    & \multicolumn{4}{c}{Motion} & \multicolumn{2}{c}{Transition (30 frames)} \\
    \hline 
    & R-precision $\uparrow$ & FID $\downarrow$ & Diversity $\rightarrow$ & MultiModal-Dist $\downarrow$ & FID $\downarrow$ & Diversity $\rightarrow$ \\
    \hline
    Ground Truth & $0.797 \pm 0.003$ & $1.6 \cdot 10^{-3} \pm 0.00$ & $9.59 \pm 0.13$ & $2.98 \pm 0.01$ & $1.8 \cdot 10^{-3} \pm 0.00$ & $9.55 \pm 0.09$ \\
    \hline 

    $\mathbf{W} = 30, P=30 $ & $\mathbf{0.681 \pm 0.011}$ & $\mathbf{0.41 \pm 0.09}$ & $\mathbf{9.60 \pm 0.14}$ & $\underline{3.18 \pm 0.01}$ & $4.33 \pm 0.31$ & $7.62 \pm 0.09$ \\
    $\mathbf{W} = 90, P=30$ & $0.674 \pm 0.012$ & $0.48 \pm 0.09$&  ${9.63 \pm 0.16}$ & $3.20 \pm 0.01$ & $\underline{2.47 \pm 0.30}$ & $\underline{8.47 \pm 0.07}$\\
    $\mathbf{W} = 120, P=30$ & ${0.679 \pm 0.007}$ & ${0.47 \pm 0.12}$&  $\underline{9.58 \pm 0.15}$ & $\mathbf{3.15 \pm 0.01}$ & $\mathbf{2.04 \pm 0.30}$ & $\mathbf{8.69 \pm 0.09}$\\
    $\mathbf{W} = 150, P=30$ & $0.667 \pm 0.014$ & $0.53 \pm 0.09$&  $9.46 \pm 0.14$ & $3.37 \pm 0.01$ & $2.67 \pm 0.35$ & $8.41 \pm 0.09$\\
    \hline
    $W = 120, \mathbf{P} = 40$ & $0.677 \pm 0.010$ & $0.47 \pm 0.09$&  $9.56 \pm 0.17$ & $3.38 \pm 0.01$ & $2.53 \pm 0.33$ & $8.40 \pm 0.09$\\
    $W = 120, \mathbf{P} = 80$ & $0.677 \pm 0.007$ & $0.47 \pm 0.07$&  ${9.54 \pm 0.09}$ & ${3.31 \pm 0.01}$ & $3.07 \pm 0.33$ & $8.15 \pm 0.15$\\
    $W = 120, \mathbf{P} = 120$ & $\underline{0.680 \pm 0.010}$ & ${0.46 \pm 0.09}$ &  $9.61 \pm 0.07$ & $3.25 \pm 0.01$ & $4.70 \pm 0.32$ & $6.87 \pm 0.25$ \\
    $W = 120, \mathbf{P} = 160$ & ${0.678 \pm 0.011}$ & $\underline{0.43 \pm 0.07}$ &  $9.58 \pm 0.08$ & $3.24 \pm 0.01$ & $7.47 \pm 0.39$ & $6.32 \pm 0.53$ \\

    \hline
    \bottomrule 
    
\end{tabular}%
}
\vspace{-2.5mm}
\caption{Ablation Study on Sliding Window Size $W$ and Stride Size $P$. Experimental results show that as $W$ increases, the alignment between individual motions and text decreases due to the addition of more contextual information. When $P \geq W$, some motion segments cannot be sampled, which has a negative impact on randomly initialized transition segments.}
\label{tab:abl_patch}

\end{table*}


%% file: sec/5_conclusion.tex
\section{Conclusion}
In this paper, we present InfiniDreamer, a novel framework for generating arbitrarily long motion sequences from a list of textual descriptions. InfiniDreamer models the entire sequence as a set of differentiable parameters and optimizes them via our proposed Segment Score Distillation (SSD), a method that leverages a pre-trained motion diffusion prior to guide short motion segments. Specifically, we iteratively sample short motion segments from a randomly initialized long-sequence motion, and optimize each to align with the diffusion prior distribution. This iterative process ultimately yields a coherent long-sequence motion. Our experiments validate that InfiniDreamer possesses exceptional contextual understanding, enabling it to generate natural and smooth transitions, thereby synthesizing arbitrarily long human motion sequence. Notably, our approach is independent of any specific motion diffusion prior. In other words, short motion clips generation and the ensemble of long motion sequences are decoupled. Therefore, future advancements in diffusion models for short motion generation could further enhance the performance InfiniDreamer. We hope our approach inspires new directions in long-sequence motion generation.

\section*{Acknowledgments}
This work is supported by National Natural Science Foundation of China (U2336212), National Natural Science Foundation of China (62472381), Fundamental Research Funds for the Zhejiang Provincial Universities (226-2024-00208). Earth System Big Data Platform of the School of Earth Sciences, Zhejiang University.


%% file: sec/X_suppl.tex
\clearpage
\setcounter{page}{1}
\maketitlesupplementary

\section{Theoretical Analysis: Segment Score Distillation and Global Consistency}

We provide a theoretical analysis to demonstrate that if the Segment Score Distillation loss 
$\mathcal{L}_{\text{align}} = \mathbb{E}_{t, \epsilon}\big[w(t)\|\hat{x}^i_0 - x^i_0\|^2_2\big]$ converges, the resulting long motion sequence M is guaranteed to be globally coherent and smooth.

\subsection{Problem Setup}

Let $M = \{m_1, t_1, m_2, t_2, ..., m_N\}$ denote a long motion sequence $M$, where $m_i$ represents the motion segment corresponding to the $i$-th text prompt, and $t_i$ represents the transition segment between $m_i$ and $m_{i+1}$. The initial sequence $M$ is constructed by concatenating motion segments $\{m_i\}$ with randomly initialized transition $\{t_i\}$. Our goal is to iteratively optimize $M$ so that:

\noindent (i) Each motion segment $m_i$ and transition $t_i$ conforms to a learned motion prior $p(x)$, ensuring realism.

\noindent (ii) $M$ achieves global coherence and smoothness.

To optimize $M$, we introduce Segment Score Distillation (SSD), which operates as follows:

\noindent (i) Using a sliding window, we sample overlapping short sequences $\{x^i_0\}^K_{i=1}$ from $M$, where each $x^i_0 = M[i: i+L]$ spans motion segments $(m_i, m_{i+1}$ and transitions $(t_i)$.

\noindent (ii) Add noise to each sampled sequence to obtain $x^i_t \sim q(x^i_t|x^i_0)$.

\noindent (iii) Denoise $x^i_t$ using the Motion Diffusion Model $\phi$ to predict $\hat{x}^i_0$, and compute the alignment loss:

\begin{equation}
    \label{eq:align_loss_supp}
    \begin{aligned}
        \mathcal{L}_{align} = \mathbf{E}_{t, \epsilon}[w(t)||\hat{x}^i_0 - x^i_0||^2_2], 
    \end{aligned}
\end{equation}

\noindent (iv) The loss $\mathcal{L}_{align}$ is back-propagated to optimize $M$, ensuring that both $m_i$ and $t_i$ align with $p(x)$

\subsection{Theoretical Analysis}

We now prove that minimizing $\mathcal{L}_{align}$ ensures global coherence and smoothness in M.

\subsubsection{Local Consistency Through Loss Convergence}

The Motion Diffusion Model $\phi$ is trained to model the conditional distribution of motion $x^i_0$ at time step $t$:

\begin{equation}
    \label{eq:distribution}
    \begin{aligned}
        \hat{x}^i_0 = \phi(x^i_t, t), \;\;x^i_t \sim q(x^i_t|x^i_0),
    \end{aligned}
\end{equation}
where $q(x^i_t|x^i_0)$ represents the forward diffusion process.

When $\mathcal{L}_{align} \rightarrow 0$, the predicted $\hat{x}^i_0$ aligns with the $x^i_0$ for all sampled segments. This implies:

\noindent (i) Each segment $x^i_0$ conforms to the motion prior $p(x)$, which encodes realistic and smooth dynamics.

\noindent (ii) The transitions within $x^i_0$ are temporally consistent.

\subsubsection{Global Coherence via Overlapping Optimization}

The sliding window mechanism ensures that adjacent sampled sequences overlap. Let:

\begin{equation}
    \label{eq:overlap_seg}
    \begin{aligned}
        x^i_0 = M[i: i+L], \;\; x^{i+1}_0 = M[i+\delta: i+L + \delta],
    \end{aligned}
\end{equation}
where $L$ is the segment length, and $\delta$ is the overlap step size. The overlapping region $O^i = x^i_0 \bigcap x^{i+1}_0$ satisfies:

\begin{equation}
    \label{eq:region_satis}
    \begin{aligned}
        \hat{x}^i_0[O^i] \rightarrow x^i_0[O^i], \;\; \hat{x}^{i+1}_0[O^i] \rightarrow x^{i+1}_0[O^i], 
    \end{aligned}
\end{equation}
by transitivity:

\begin{equation}
    \label{eq:transitivity}
    \begin{aligned}
        ||\hat{x}^i_0[O^i] - \hat{x}^{i+1}_0[O^i]||^2_2 \rightarrow 0,
    \end{aligned}
\end{equation}
as $\mathcal{L}_{align} \rightarrow 0$. By enforcing consistency within overlapping regions, the optimization propagates coherence across the entire sequence M. 

\subsubsection{Smoothness Guarantee}

Smoothness in $M$ is ensured through two mechanisms:

(1) \textbf{Local Smoothness}: The loss $\mathcal{L}_{align}$ minimizes the discrepancy between $\hat{x}^i_0$ and $x^i_0$, ensuring smooth dynamics within each short segment.

(2) \textbf{Global Smoothness}: Overlapping regions $O^i$ propagate smoothness across segment boundaries.

By the end of optimization, all segments and transitions are aligned with $p(x)$, resulting in a globally smooth sequence $M$.

\section{Further implementation details}
To facilitate better reproducibility of our work, we provide additional details about our implementation in this section. For HumanML3D, we set the fps as $20$, and encode timesteps as a sinusoidal positional encoding. We utilize a dense layer to encode poses of 263D into a sequence of 512D vectors. For BABEL, we set fps as $30$. We encode poses of 135D into a sequence of 512D vectors. In the first stage, we utilize guidance scale $2.5$ to generate each single motion segments, and in the process of Segment Score Distillation, we utilize $7.5$ to optimize the entire motion sequence. We set the weighting function $w(t)$ as $1-\alpha(t)$ for all experiments.

\input{table/supp_humanml}

\input{table/supp_babel}

\section{More experimental results}
In this section, we present more experimental results to validate the effectiveness of our framework.

\subsection{More quantitative results}
We also compare our framework with the latest work, FlowMDM~\cite{barquero2024seamless}, which introduces Blended Positional Encodings, a technique that combines absolute and relative positional encodings in the denoising process. We follow the evaluation protocol of FlowMDM~\cite{barquero2024seamless}. However, we only use FlowMDM~\cite{barquero2024seamless} to generate individual motion segments, and then use our method to generate the entire long motion sequence. As shown in Tab.~\ref{tab:supp_humanml} and Tab.~\ref{tab:supp_babel}, we find that InfiniDreamer performs slightly worse than FlowMDM~\cite{barquero2024seamless} but outperforms previous training-free methods. We speculate that this is because FlowMDM~\cite{barquero2024seamless} is fine-tuned on long human motion sequences using both absolute and relative positional encodings, which introduces some interference in individual short motion segments. Our method, which adds further interference, therefore achieves slightly lower performance compared to FlowMDM~\cite{barquero2024seamless}. Nonetheless, the experimental results demonstrate the advantages of our approach over DoubleTake~\cite{shafir2023human}.

\input{fig/supp_compare}
\input{table/abl_prompt}
\input{fig/single_long}

\subsection{More qualitative results}
We also conducted qualitative experiments to compare the results of our framework with those of FlowMDM~\cite{barquero2024seamless}. In this section, we use the open-source model of FlowMDM~\cite{barquero2024seamless} to generate its results, while for our method, we use MDM~\cite{tevet2023human} as our motion prior. As shown in Fig.~\ref{fig:supp_compre}, we present two examples of long motion sequence generation. The first example, at the top of the figure, is generated using the text prompts `jogging forward slowly' + `a person is walking down the stairs' + `jogging forward slowly'. The results show that both FlowMDM and our method can infer the transitional `walking up the stairs' segment before descending. However, FlowMDM exhibits slight motion drift during this segment. In the second example, with the text prompts `a person is walking straight' + `side steps' + `he is walking backward', we observe that FlowMDM generates the `side steps' motion incorrectly and also shows motion drift at the end. In contrast, our method avoids these issues, producing more accurate and coherent results. Additionally, we observe that the motions generated by FlowMDM exhibit a larger displacement range, while our method produces smoother and more controlled movements.

\subsection{More ablation study on prompts}
In this section, we present additional ablation studies. We explore the use of different text conditions to guide the optimization of Segment Score Distillation (SSD). In this experiments, we remove the original text selection strategy and instead optimize using a single text prompt. We present two types of prompts: ``transition'' and ``motion '', and set the guidance scale as $7.5$. We also present an unconditional optimization scenario. The results are shown in Tab.~\ref{tab:abl_prompt}, we observe that incorporating suboptimial prompts negatively impacts InfiniDreamer's performance. Using the unconditional optimization results in a slight performance decline, whereas using ``transition'' or ``motion'' as prompts leads to a more significant degradation in performance. We believe this is because the chosen text prompts are struggle to capture the semantic diversity of various transition segments.

\subsection{Long motion generation with Single Prompt}
Our framework has an additional capability: it can generate long motion sequences from a single text prompt. It is a feature that currently beyond the reach of other models. Specifically, given a short-sequence generation model, Motion Diffusion Model (MDM)~\cite{tevet2023human}, we set the total frame count of the long sequence to 520 frames and the frame count of each short sequence to 120 frames. We employ conditional Segment Score Distillation (SSD), using the text prompt as the conditioning input. At the beginning, we randomly initialize a long motion sequence. In this experiment, we omit the first stage of InfiniDreamer, meaning that we do not use MDM to generate the initial short motion sequence. In the subsequent stages, we set the guidance scale to $10.0$ and the learning rate to $0.005$. As shown in Fig.~\ref{fig:single_long}, we use ``a person takes 3 steps backward'' as our textual prompt. InfiniDreamer, through conditional optimization, extends the generation capability of the original Motion Diffusion Model (MDM) from 70-200 frames to 520 frames, while maintaining alignment between the generated motions and the input text.


\section{Limitation}

InfiniDreamer is capable of generating arbitrarily long motion sequences based on text prompts, even in single-text scenarios. However, our framework still has some limitations. For example, the generation of sub-motions is constrained by the performance of the short-sequence generation model. Additionally, our method is slower compared to other sampling approaches, taking approximately 4 minutes to generate a 520-frame sequence. In the future, we plan to improve its efficiency and enhance InfiniDreamer's performance by advancing the capabilities of the short-sequence generation model.

\vspace{-1mm}
\begin{algorithm}[!t]
    \caption{Segment Score Distillation (SSD)}
    \label{alg:ssd}
    \begin{algorithmic}[1]
    \REQUIRE Initial motion sequence $M_0$, motion diffusion model $\phi$, number of iteration $N$, windows size $W$, stride $S$, learning rate $\eta$
    \ENSURE Optimized long motion sequence $M$
        
    \STATE \textbf{Initialize:} $M \gets M_0$
    \STATE \textbf{Set:} Gradient masks  
    \FOR{$n=1$ to $N$}
        \STATE Sample a start index $i$ from $\{1, 1+S, \dots, \text{len}(M) - W\}$
        \STATE Extract a motion segment $x^i_0 \gets M[i : i+W]$
        \STATE Choose the textual prompt based on strategy
        \STATE Sample a timestep $t \sim \mathcal{U}(20, 980)$
        \STATE Add noise to obtain $x^i_t = \sqrt{\alpha_t} x^i_0 + \sqrt{\sigma_t} \epsilon$, where $\epsilon \sim \mathcal{N}(0, I)$
        \STATE Compute total SSD loss via Eq.~\ref{eq:slide_score_distillation}
        \STATE Update $M$ by backpropagating $\mathcal{L}_{ssd}$ and adjusting the values of $x^i_0$ in $M$
    \ENDFOR
    \STATE \RETURN Optimized long motion sequence $M$
    \end{algorithmic}
\end{algorithm}

%% file: table/supp_humanml.tex
\begin{table*}[!t]
\centering
\setlength{\tabcolsep}{4pt}
\renewcommand{\arraystretch}{1.2}
\resizebox{\textwidth}{!}{%
\begin{tabular}{lcccc||cccc}
    \toprule 
    & \multicolumn{4}{c}{Motion} & \multicolumn{4}{c}{Transition} \\
    \hline 
    & R-prec $\uparrow$ & FID $\downarrow$ & Div $\rightarrow$ & MM-Dist $\downarrow$ & FID $\downarrow$ & Div $\rightarrow$ &PJ $\rightarrow$ & AUJ $\downarrow$\\
    \hline
    Ground Truth & $0.796 \pm 0.004$ & $0.00 \pm 0.00$ & $9.34 \pm 0.08$ & $2.97 \pm 0.01$ & $0.00 \pm 0.00$ & $9.54 \pm 0.15$ & $0.02 \pm 0.00$ & $0.00 \pm 0.00$ \\
    \hline 
    DoubleTake*  & ${0.643}^{\pm0.005}$ & ${0.80}^{\pm0.02}$ & ${9.20}^{\pm0.11}$ & ${3.92}^{\pm0.01}$  & ${1.71}^{\pm0.05}$ & $\underline{8.82}^{\pm0.13}$ & $0.52^{\pm0.01}$ & $2.10^{\pm0.03}$  \\
    DoubleTake  & $0.628^{\pm0.005}$ & $1.25^{\pm0.04}$ & $9.09^{\pm0.12}$ & $4.01^{\pm0.01}$  & $4.19^{\pm0.09}$ & $8.45^{\pm0.09}$ & $0.48^{\pm0.00}$ & $1.83^{\pm0.02}$  \\
    MultiDiffusion  & $0.629^{\pm0.002}$ & $1.19^{\pm0.03}$ & $\textbf{9.38}^{\pm0.08}$ & $4.02^{\pm0.01}$  & $4.31^{\pm0.06}$ & $8.37^{\pm0.10}$ & $\underline{0.17}^{\pm0.00}$ & $\underline{1.06}^{\pm0.01}$  \\
    DiffCollage  & $0.615^{\pm0.005}$ & $1.56^{\pm0.04}$ & $8.79^{\pm0.08}$ & $4.13^{\pm0.02}$  & $4.59^{\pm0.10}$ & $8.22^{\pm0.11}$ & $0.26^{\pm0.00}$ & $2.85^{\pm0.09}$  \\
    FlowMDM  & $\underline{0.685}^{\pm0.004}$ & $\textbf{0.29}^{\pm0.01}$ & $9.58^{\pm0.12}$ & ${3.61}^{\pm0.01}$  & $\textbf{1.38}^{\pm0.05}$ & ${8.79}^{\pm0.09}$ & $\textbf{0.06}^{\pm0.00}$ & $\textbf{0.51}^{\pm0.01}$  \\
    \hline
    InfiniDreamer (ours) & ${0.679} ^{\pm 0.007}$ & $\underline{0.47} ^ {\pm 0.12}$ & ${9.58} ^{\pm 0.15}$ & $\mathbf{3.15} ^{\pm 0.01}$ & ${2.04}^{\pm 0.05}$ & $8.69^{\pm 0.11}$ & - & - \\
    \quad + FlowMDM~\cite{barquero2024seamless} & $ 0.674 ^{\pm 0.004}$ & ${0.68} ^ {\pm 0.02}$ & ${9.27} ^{\pm 0.11}$ & ${3.78} ^{\pm 0.01}$ & ${1.64}^{\pm 0.05}$ & ${8.77}^{\pm 0.11}$ & - & - \\

    \quad + MLD~\cite{chen2023executing} & $\mathbf{0.713} ^{\pm 0.003}$ & ${0.52} ^ {\pm 0.15}$ & $\underline{9.46} ^{\pm 0.11}$ & $\underline{3.17} ^{\pm 0.01}$ & $\underline{1.47}^{\pm 0.05}$ & $\mathbf{8.87}^{\pm 0.11}$ & - & - \\

    \bottomrule 
    
\end{tabular}%
}
\vspace{-1mm}
\caption{Comparison of InfiniDreamer with the state of the art in HumanML3D. Symbols $\uparrow$, $\downarrow$, and $\rightarrow$ mean that higher, lower, or closer to the ground truth (GT) value are better, respectively. We run each evaluation 10 times to obtain the final results. We use \textbf{Bold} to indicate the best result, and use \underline{underline} to indicate the second-best result.  }
\label{tab:supp_humanml}
\end{table*}

%% file: table/supp_babel.tex
\begin{table*}[!t]
\centering
\setlength{\tabcolsep}{4pt}
\renewcommand{\arraystretch}{1.2}
\resizebox{\textwidth}{!}{%
\begin{tabular}{lcccc||cccc}
    \toprule 
    & \multicolumn{4}{c}{Motion} & \multicolumn{4}{c}{Transition} \\
    \hline 
    & R-prec $\uparrow$ & FID $\downarrow$ & Div $\rightarrow$ & MM-Dist $\downarrow$ & FID $\downarrow$ & Div $\rightarrow$ &PJ $\rightarrow$ & AUJ $\downarrow$\\
    \hline
    Ground Truth & $0.715^{\pm0.003}$ & $0.00^{\pm0.00}$ & $8.42^{\pm0.15}$ & $3.36^{\pm0.00}$  & $0.00^{\pm0.00}$ & $6.20^{\pm0.06}$ & $0.02^{\pm0.00}$ & $0.00^{\pm0.00}$ \\
    \hline 
    TEACH\_B  & $\textbf{0.703}^{\pm0.002}$ & $1.71^{\pm0.03}$ & $8.18^{\pm0.14}$ & $\textbf{3.43}^{\pm0.01}$  & ${3.01}^{\pm0.04}$ & $\textbf{6.23}^{\pm0.05}$ & $1.09^{\pm0.00}$ & $2.35^{\pm0.01}$  \\
    TEACH  & $0.655^{\pm0.002}$ & $1.82^{\pm0.02}$ & $7.96^{\pm0.11}$ & $3.72^{\pm0.01}$  & $3.27^{\pm0.04}$ & $6.14^{\pm0.06}$ & $\underline{0.07}^{\pm0.00}$ & $\underline{0.44}^{\pm0.00}$  \\
    DoubleTake*  & $0.596^{\pm0.005}$ & $3.16^{\pm0.06}$ & $7.53^{\pm0.11}$ & $4.17^{\pm0.02}$  & $3.33^{\pm0.06}$ & $\underline{6.16}^{\pm0.05}$ & $0.28^{\pm0.00}$ & $1.04^{\pm0.01}$  \\
    DoubleTake  & $0.668^{\pm0.005}$ & ${1.33}^{\pm0.04}$ & $7.98^{\pm0.12}$ & $3.67^{\pm0.03}$  & $3.15^{\pm0.05}$ & $6.14^{\pm0.07}$ & $0.17^{\pm0.00}$ & $0.64^{\pm0.01}$  \\
    MultiDiffusion  & $\underline{0.702}^{\pm0.005}$ & $1.74^{\pm0.04}$ & $\textbf{8.37}^{\pm0.13}$ & $\textbf{3.43}^{\pm0.02}$  & $6.56^{\pm0.12}$ & $5.72^{\pm0.07}$ & $0.18^{\pm0.00}$ & $0.68^{\pm0.00}$  \\
    DiffCollage  & $0.671^{\pm0.003}$ & $1.45^{\pm0.05}$ & $7.93^{\pm0.09}$ & $3.71^{\pm0.01}$  & $4.36^{\pm0.09}$ & $6.09^{\pm0.08}$ & $0.19^{\pm0.00}$ & $0.84^{\pm0.01}$  \\
    FlowMDM  & $\underline{0.702}^{\pm0.004}$ & $\underline{0.99}^{\pm0.04}$ & $\underline{8.36}^{\pm0.13}$ & $3.45^{\pm0.02}$  & $\underline{2.61}^{\pm0.06}$ & $6.47^{\pm0.05}$ & $\textbf{0.06}^{\pm0.00}$ & $\textbf{0.13}^{\pm0.00}$ \\
    \hline
    InfiniDreamer (ours) & $0.543 ^{\pm 0.009}$ & $\textbf{0.97} ^{\pm 0.09}$ & $8.31 ^{\pm 0.06} $ & $5.80 ^{\pm 0.01}$ & $\textbf{2.07} ^{\pm 0.30}$ & $7.95 ^{\pm 0.07}$ & - & - \\
    
    \quad + FlowMDM~\cite{barquero2024seamless} & ${0.667} ^{\pm 0.005}$ & $1.49 ^ {\pm 0.03}$ & ${7.94} ^{\pm 0.14}$ & ${3.53} ^{\pm 0.01}$ & ${2.97}^{\pm 0.08}$ & $6.31^{\pm 0.07}$ & - & - \\

    \bottomrule 
    
\end{tabular}%
}
\vspace{-1mm}
\caption{Comparison of InfiniDreamer with the state of the art in BABEL. Symbols $\uparrow$, $\downarrow$, and $\rightarrow$ mean that higher, lower, or closer to the ground truth (GT) value are better, respectively. We run each evaluation 10 times to obtain the final results. We use \textbf{Bold} to indicate the best result, and use \underline{underline} to indicate the second-best result.  }
\label{tab:supp_babel}
\end{table*}

%% file: fig/supp_compare.tex
\begin{figure*}[!ht]
  \centering
  \includegraphics[width=0.98\textwidth]{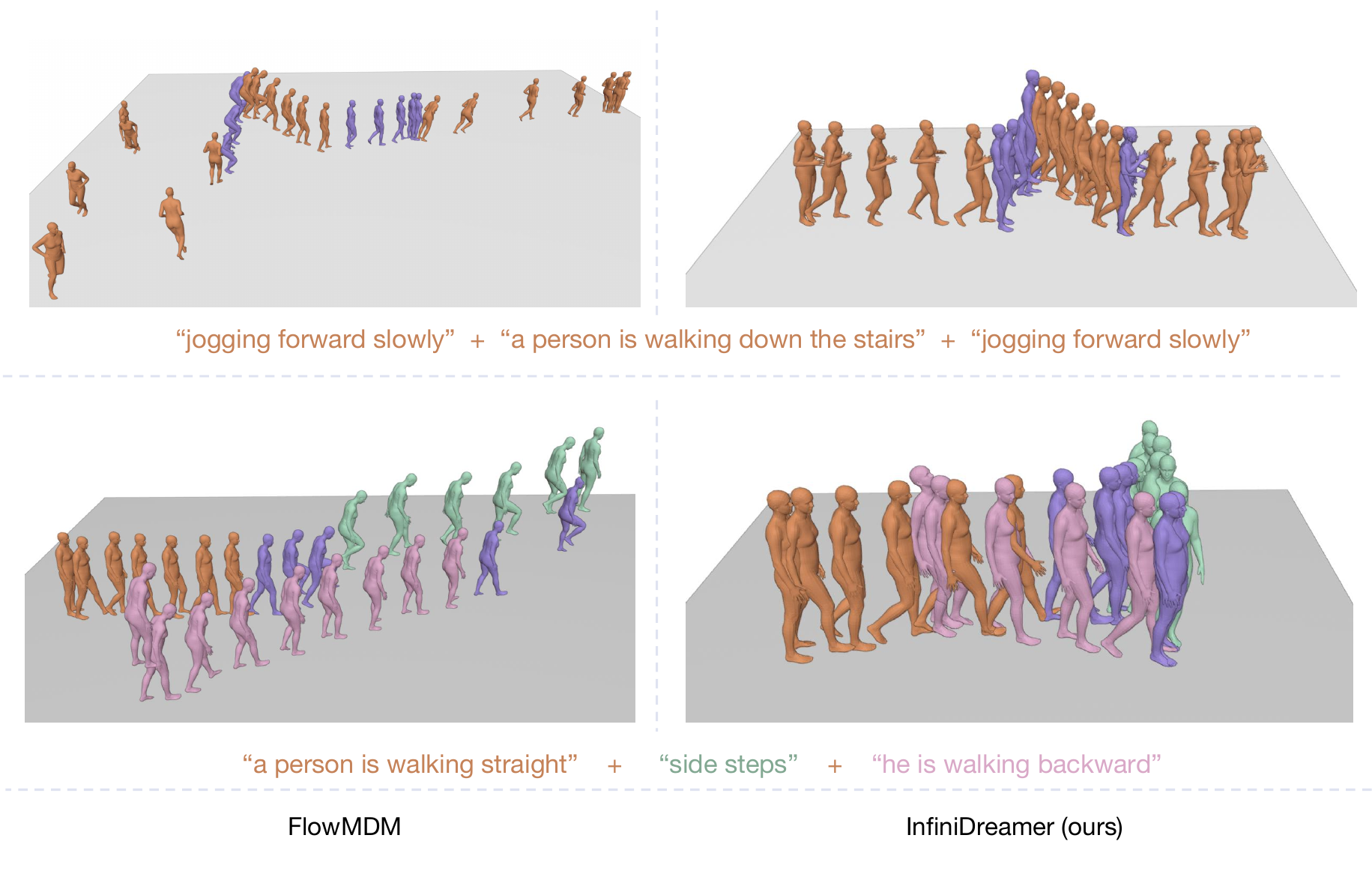}
  \vspace{-6mm}
  \caption{Qualitative Comparisons to FlowMDM for Long Motion Generation. We present two examples: in the top row, our framework demonstrates strong contextual understanding, guiding the transition segment to ``go upstairs" in response to the following ``downstairs" prompt. In contrast, FlowMDM shows slightly motion drift in this segment. In the bottom row, we use a more fine-grained textual prompt, where the FlowMDM exhibits issues with motion drift and semantic errors, failing to generate the ``side steps'' segment. Our framework, however, produces a higher-quality sequence with enhanced fine-grained comprehension of the text.}
\label{fig:supp_compre}
\end{figure*}

%% file: table/abl_prompt.tex
\begin{table*}[!t]
\centering
\setlength{\tabcolsep}{4pt}
\renewcommand{\arraystretch}{1.2}
\resizebox{\textwidth}{!}{%
\begin{tabular}{lcccc||cc}
    \toprule 
    & \multicolumn{4}{c}{Motion} & \multicolumn{2}{c}{Transition (30 frames)} \\
    \hline 
    & R-precision $\uparrow$ & FID $\downarrow$ & Diversity $\rightarrow$ & MultiModal-Dist $\downarrow$ & FID $\downarrow$ & Diversity $\rightarrow$ \\
    \hline
    Ground Truth & $0.797 \pm 0.003$ & $1.6 \cdot 10^{-3} \pm 0.00$ & $9.59 \pm 0.13$ & $2.98 \pm 0.01$ & $1.8 \cdot 10^{-3} \pm 0.00$ & $9.55 \pm 0.09$ \\
    \hline 
    InfiniDreamer (ours) & $\mathbf{0.679 \pm 0.007}$ & $\mathbf{0.47 \pm 0.12}$ & $\mathbf{9.58 \pm 0.15} $ & $\mathbf{3.15 \pm 0.01}$ & $\mathbf{2.04\pm 0.28}$ & $\mathbf{8.69 \pm 0.09}$\\  
    \hline 
    w/o text selection strategy: \\
    
    \quad + ``transition'' & ${0.657 \pm 0.011}$ & ${0.53 \pm 0.13}$ & ${9.51 \pm 0.13} $ & ${3.32 \pm 0.01}$ & ${2.32\pm 0.32}$ & ${8.43 \pm 0.09}$\\   
    \quad + ``motion'' & ${0.650 \pm 0.013}$ & ${0.56 \pm 0.14}$ & ${9.54 \pm 0.13} $ & ${3.39 \pm 0.01}$ & ${2.36\pm 0.33}$ & ${8.46 \pm 0.10}$\\    
    \quad + uncondition & $\underline{0.664 \pm 0.010}$ & $\underline{0.49 \pm 0.09}$ & $\underline{9.55 \pm 0.13} $ & $\underline{3.23 \pm 0.01}$ & $\underline{2.15\pm 0.28}$ & $\underline{8.57 \pm 0.11}$\\    
    \hline
    \bottomrule 
    
\end{tabular}%
}
\vspace{-2.5mm}
\caption{Ablation Study on textual prompt. We remove the original text selection strategy and instead optimize using a single text prompt. We present two types of prompt: ``transition'' and ``motion'', as well as an unconditional optimization scenario. We find that mismatched textual conditions lead to a decline in performance, while the unconditional setting produces an sub-optimal result. We believe this is because the text prompts used are not well-suited to capture the semantics of diverse transition segments. This validate the effectiveness of our text condition selection strategy.}

\label{tab:abl_prompt}

\end{table*}


%% file: fig/single_long.tex
\begin{figure}[!ht]
  \centering
  \includegraphics[width=\columnwidth]{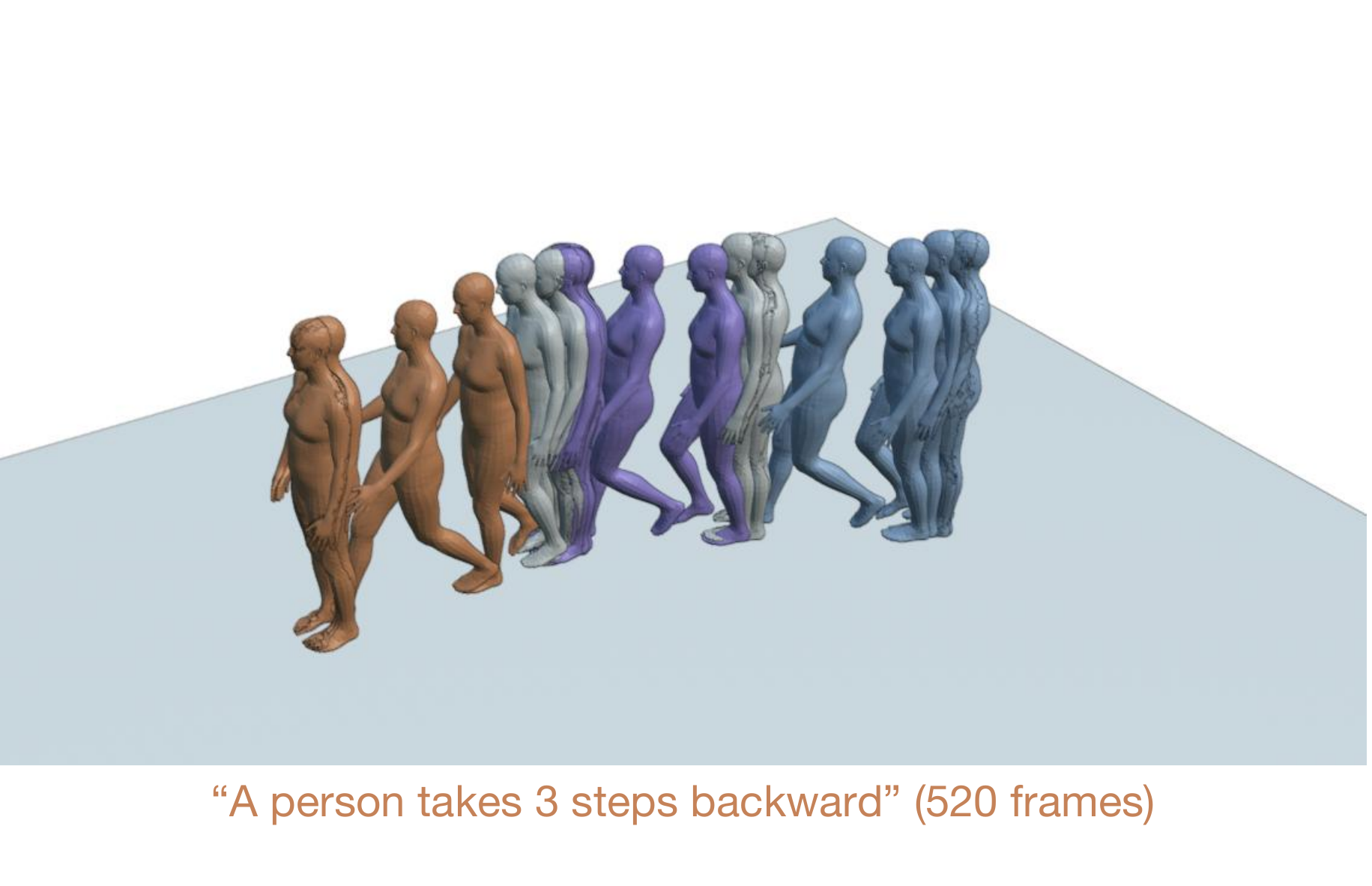}
  \vspace{-8mm}
  \caption{We demonstrate InfiniDreamer's capability to generate long motion sequence from a single text prompt. Starting with a base model designed to generate short sequences (approximately 70 to 200 frames), our framework extends its generation range to 520 frames while ensuring that the generated motions remain semantically consistent with the input text.}
\label{fig:single_long}
\end{figure}